\documentclass[10pt,twocolumn,letterpaper]{article}

\usepackage{iccv}
\usepackage{times}
\usepackage{epsfig}
\usepackage{graphicx}
\usepackage{amsmath}
\usepackage{amssymb}
\usepackage{comment}

\usepackage[font=small,labelfont=bf,tableposition=top]{caption}
\usepackage[pagebackref=true,breaklinks=true,letterpaper=true,colorlinks,bookmarks=false]{hyperref}
\usepackage{subfigure}
\usepackage{multirow}
\usepackage{booktabs}
\newcommand{\hsl}[1]{#1}
\newcommand{\hslc}[1]{}

\newcommand{\hoe}[1]{#1}

\iccvfinalcopy 

\def\iccvPaperID{2687} 

\ificcvfinal\fi

\begin{document}

\title{\hsl{HMC: Hierarchical Mesh Coarsening for Skeleton-free Motion Retargeting}}

\author{Haoyu Wang\\
Tsinghua University\\
Shenzhen, China\\
{\tt\small wanghaoy21@mails.tsinghua.edu.cn}
\and
Shaoli Huang\\
Tencent AI Lab\\
Shenzhen, China\\
{\tt\small shaol.huang@gmail.com}
\and
Fang Zhao\\
Tencent AI Lab\\
Shenzhen, China\\
{\tt\small zhaofang0627@gmail.com}
\and
Chun Yuan\\
Tsinghua university\\
Shenzhen, China\\
{\tt\small yuanc@sz.tsinghua.edu.cn}
\and
Ying Shan\\
Tencent\\
Shenzhen, China\\
{\tt\small yingsshan@tencent.com}
}

\ificcvfinal\thispagestyle{empty}\fi

\twocolumn[{
\maketitle
\begin{center}
    \captionsetup{type=figure}
    \vspace{-2em}
    \includegraphics[width=0.96\textwidth]{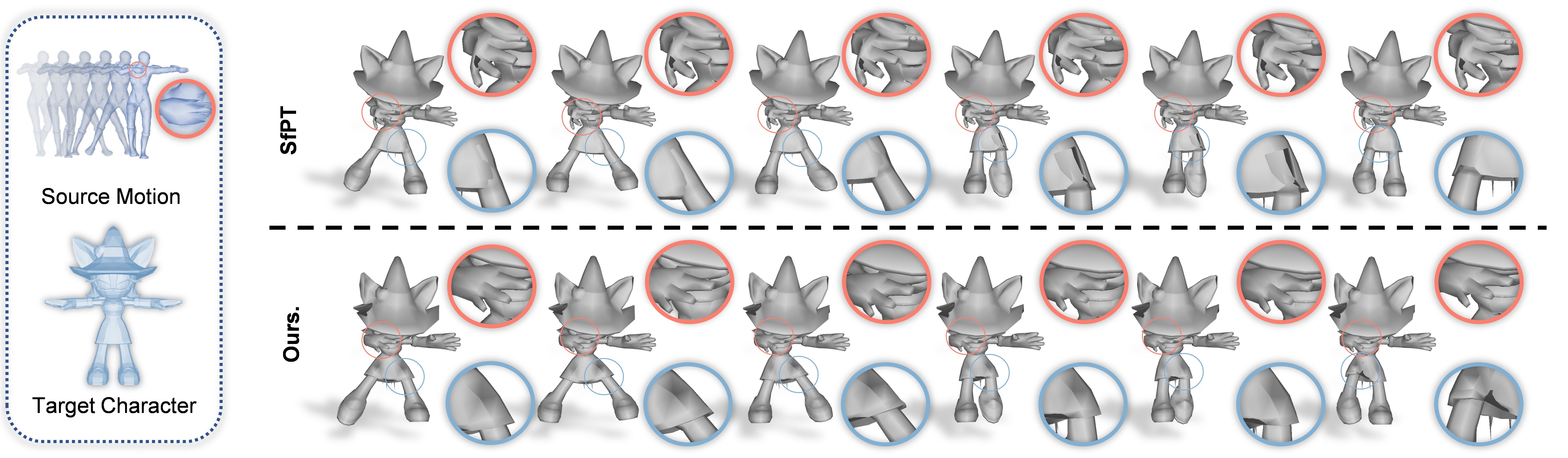}
    \vspace{-1em}
    \captionof{figure}{Skeleton-free motion retargeting from an input source motion to the target character. \textbf{Left: } Source motion is animated on a Mixamo~\cite{mixamo2023} character, and the target character is selected in RigNet~\cite{xu2020rignet} dataset. The skeletal rigging and skinning on both of the input characters are agnostic during inference. 
    \textbf{Right: } Comparison between SfPT~\cite{liao2022skeleton} (the state-of-art method) and ours shows the superior performance of our model on \hoe{handling small-part motions and preserving local motion interdependence.} Thereinto, \textcolor[RGB]{250,127,111}{red bubble} reflects \hoe{the consistency of the right hand's direction between the source and target characters}.
    \textcolor[RGB]{130,176,210}{blue bubble} reflects \hoe{local motion independence between the leg and skirt (i.e., whether leg-skirt interpenetration occurs).} }
    \label{fig::title}
\end{center}
}]

\begin{abstract}
    \hsl{We present a simple yet effective method for skeleton-free motion retargeting.}
    \hsl{Previous methods transfer motion between high-resolution meshes, failing to preserve the inherent local-part motions in the mesh.}
    Addressing this issue, our proposed method learns the correspondence in a coarse-to-fine fashion by integrating the retargeting process with a mesh-coarsening pipeline. \hsl{First, we propose a mesh-coarsening module that coarsens the mesh representations for \hoe{better} motion transfer. This module improves the ability to handle small-part motion and preserves the local motion interdependence between neighboring mesh vertices.
    Furthermore, we leverage a hierarchical refinement procedure to complement missing mesh details by gradually improving the low-resolution mesh output with a higher-resolution one.}
    \hsl{We evaluate our method on several well-known 3D character datasets, and it} yields an average improvement of 25\% on point-wise mesh euclidean distance (PMD) against the start-of-art method.
     \hsl{Moreover, our qualitative results show that our method}
    is significantly helpful in preserving the moving consistency of different body parts on the target character due to disentangling body-part structures and mesh details in a hierarchical way. 
\end{abstract}


\section{Introduction\label{sec::intro}}
Posing and animating various 3D characters~\cite{weng2019photo,yuan2021simpoe,li2021multivisual,brodt2022sketch2pose} are critical demands in the game and animation industry, which conventionally requires backbreaking character-dependent manual processes such as skeletal rigging~\cite{Allen2011Body,arshad2019physical,sarkkomaa2022facial} and skinning~\cite{Chaudhry2010Character,komaritzan2019fast,toothman2020expressive}. 
Along with the rapid emergence of massive 3D characters nowadays, learning-based motion retargeting methods that automatically apply existing motions to newly-created characters have been increasingly considered as an alternative in the past few years.
These methods are categorized into skeleton-based~\cite{villegas2018neural,aberman2020skeleton,villegas2021contact,zhu2022mocanet} and skeleton-free~\cite{yang2018biharmonic,li2021learning,liao2022skeleton} approaches.
\hoe{Compared with skeleton-based methods that learn motion transfer based on given skeleton templates, skeleton-free methods are applicable to a widened range of 3D animation applications, owing much to the direct deformation of character meshes without skeleton prior.}
However, both of them are not without limitations when tackling enormous highly-stylized characters. 
Thereinto, skeleton-based methods still require manual skeletal rigging on new characters and fail to retarget motions between topologically dissimilar skeletons (e.g., with more arms or legs); \hoe{and existing skeleton-free methods, despite possessing stronger generalization ability on unseen characters, intuitively transfer motion between high-resolution (HR) meshes, plagued with preserving the inherent local-part motions, including handling small-part motions and preserving local motion interdependence (see examples as the \textcolor[RGB]{250,127,111}{\textbf{red}} and \textcolor[RGB]{130,176,210}{\textbf{blue}} bubbles respectively in Figure~\ref{fig::title}).}
\hoe{The former is due to an unbalanced number of HR mesh vertices on different body parts, prone to tolerating wrong motions of small parts.
The latter is because the deformation and translation of each body part are learned independently, thus overlooking their local interdependence.
}


To address these issues, we propose a simple yet effective skeleton-free motion retargeting method that learns \hoe{part-wise} correspondence in a coarse-to-fine fashion by integrating the retargeting process with a mesh-coarsening pipeline. \hoe{It improves the performance of local-part motions with two sub-techniques.} 
First, we propose a mesh-coarsening module that \hoe{coarsens the mesh representations for better motion transfer. The benefits of this module are twofold. On the one hand, it reduces and rebalances the number of vertices for body parts while keeping the morphology of the character, reaching a fair consideration of small-part motions.
On the other hand, it encloses every group of spatially adjacent vertices on the HR mesh as a whole during retargeting, thus preserving local motion interdependence between them.
}
Second, we \hoe{leverage a hierarchical refinement procedure to complement missing mesh details to achieve better mesh quality. Since learning on low-resolution (LR) meshes ignores the local relations of adjacent vertices, we gradually improve the LR mesh output by applying local deformation learned from a higher-resolution one.}
Figure~\ref{fig::title} reveals that our method \hoe{visually} outperforms the state-of-art (SOTA) method both \hoe{on small-part motion and local motion interdependency.} 

We summarize the major contributions as follows: 
i) We propose a \emph{mesh-coarsening module} to \hoe{handle small-part motion and preserve local motion interdependence between neighboring mesh vertices.}
ii) We \hoe{leverage a \emph{hierarchical refinement procedure} to complement missing mesh details and thus produce high-quality pose meshes.}
iii) We evaluate our method on several renowned 3D character datasets, and the proposed model yields an average improvement of 25\% on point-wise mesh euclidean distance (PMD) compared to the SOTA method, as well as visually \hoe{generating more encouraging results}~\footnote{To see more visualization results and animated videos, please visit our project homepage at \url{https://semanticdh.github.io/HMC/}}.



\section{Related Work\label{sec::rel}}
    \textbf{Skeleton-free Motion Retargeting~\label{sec::rel1} }In contrast with skeleton-based methods~\cite{gleicher1998retargetting,villegas2018neural,aberman2020skeleton,villegas2021contact,zhu2022mocanet} that require similar skeletal rigging and skinning on input characters, skeleton-free motion retargeting methods solely take character meshes as input and deform the target mesh as similarly as the source mesh does, applicable to a widened range of 3D animation applications. To achieve this goal without skeletons, early methods~\cite{baran2007automatic, aujay2007harmonic,poirier2009rig} automatically embed a generic skeleton structure into input characters and then solve a skeleton-based retargeting problem. Thereinto, \cite{baran2007automatic} discretizes the embedding problem by fitting a predefined skeleton with a graph that reflects the approximate medial surface of characters, and joint positions of the skeleton could be thus optimized. Unfortunately, this method relies on strong assumptions about mesh connectivity, and the predefined skeleton often fails to reflect structures of divergent meshes. Addressing this, \cite{aujay2007harmonic} proposes an expressive skeleton generation method by computing a Reeb Graph~\cite{Kunii1991Constructing} of given harmonic functions~\cite{evans1998partial}. Their work reveals that the geodesic distance function with respect to the character's head could be a good metric to determine joint positions and skinning weights. Inspired by this,\cite{poirier2009rig} further proposes a multi-resolution matching algorithm to fit the Reeb graph of a mesh with input skeleton graphs, which generates much more reasonable embedded skeletons. However, these methods tend to generate complicated character-dependent skeletons, which impacts the retargeting performance when animating a morphologically different character.
    To ease the restrictions of skeletal rigging, other works~\cite{yang2018biharmonic,yang2021multiscale,liao2022skeleton} focus on direct mesh deformation by building reliable correspondences among body parts. \cite{yang2018biharmonic} leverages a bi-harmonic weight deformation framework to produce plausible poses on target mesh when several key points on input meshes are given. Although these key points on a given source mesh could be automatically identified, this method requires an extra manual process to calibrate the corresponding points on the target mesh. \cite{yang2021multiscale} introduce a coarse-to-fine fashion to encode multi-scale mesh deformation into a latent space, and an attention module is leveraged to perceive active regions at different scales. Since the latent spaces are learned from different shapes of the same mesh, retargeting between different characters is impossible with this method. Encountering with the above issue, recently, \cite{liao2022skeleton} proposes a pioneering work that automatically transfers the deformation between different highly-stylized characters. Instead of directly supervising the model on learning correspondences between source and target characters, this method first predicts soft skinning weights of the input characters as body parts and learns rigid transformation \hoe{independently} between each pair of body parts to deform the target mesh, which could be thus applied to unseen characters when their rest poses are similar. However, this method discards the hierarchical kinematic constraints on body parts, \hoe{thus failing to preserve local motion interdependence of adjacent body parts, which also produces unnatural small-part motions due to little perception of small body parts} on target characters with a complicated mesh. In this paper, we \hoe{revisit mesh coarsening techniques and propose a mesh-coarsening module to address these issues.}


\begin{figure*}[t]
    \centering
    \includegraphics[width=1.9\columnwidth]{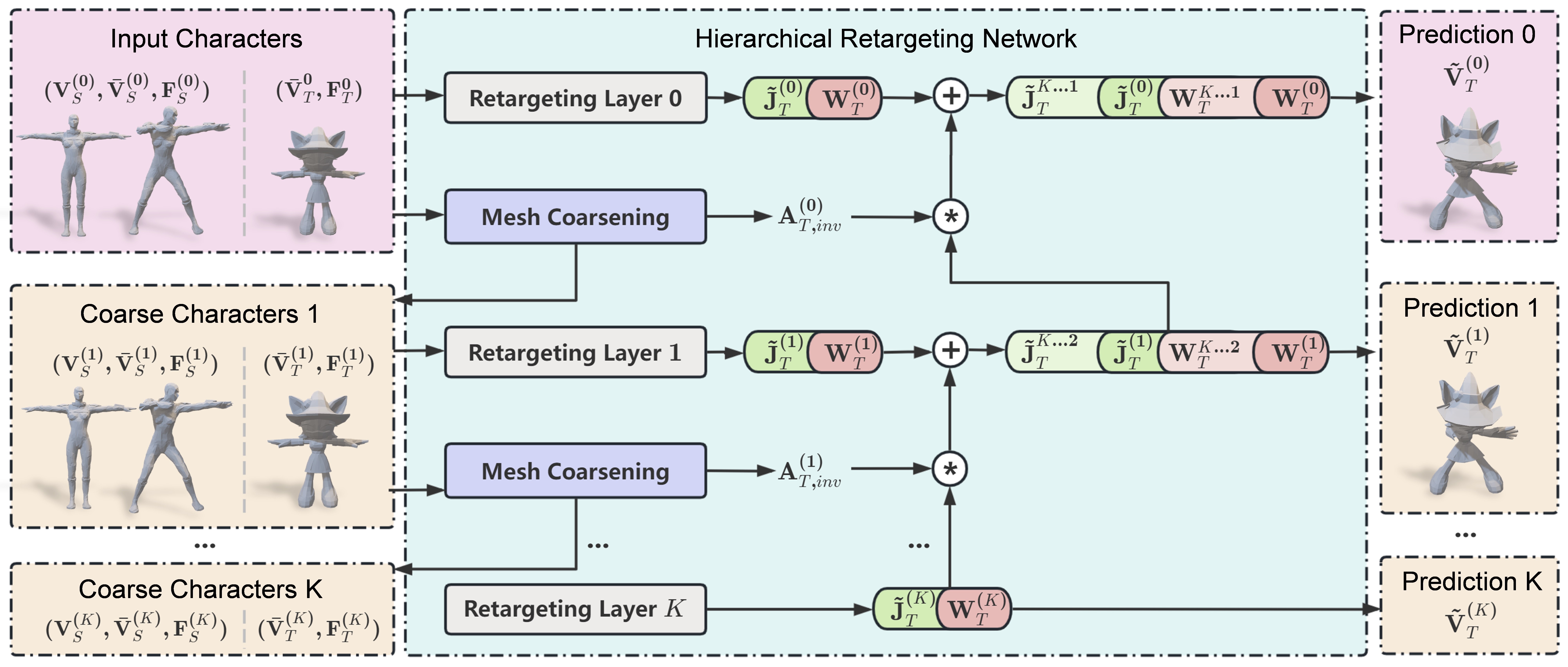}
    \caption{Overview. The network takes as input the rest and pose meshes of the source character, and the rest mesh of the target character (i.e., ${\mathbf {\bar V}}^{(0)}_S$, $\mathbf V^{(0)}_S$ and $\mathbf {\bar V}^{(0)}_T$), and then outputs the target pose mesh. \hoe{Instead of intuitively determining correspondence between the input HR meshes, we leverage a mesh-coarsening module to recursively coarsen the mesh representations for better motion transfer.} Owing much to the differentiable assignments (i.e., ${\mathbf A}^{\cdot}_{T,inv}$) constructed by the mesh-coarsening module, \hoe{the output LR pose mesh $\mathbf {\tilde V}^{(K)}_T$ can be gradually refined by applying local deformations $\{ \mathbf {J}^{(k)}_T, \mathbf W^{(k)}_T\}_{k=0}^{K-1}$ learned from higher-resolution meshes.}}
    \label{fig::pipe1}
\end{figure*}

\section{Method\label{sec::met}}
    In this section, we first introduce our retargeting framework as a generic way to retarget motions or poses among arbitrary characters without skeletons (see Section~\ref{sec::met1}). Leveraged in this framework, two proposed techniques are then described in the next Section~\ref{sec::met2} and~\ref{sec::met3}. Finally, we summarize in Section~\ref{sec::met4} the crucial losses and training strategies to reach the best performance of our model.
    
    \subsection{Overview\label{sec::met1}}
    A character mesh is formulated as a group of vertices and faces $(\mathbf V, \mathbf F)$, wherein $\mathbf V \in \mathbb R^{N\times 3}$ denotes 3d positions of $N$ vertices, and $\mathbf F \in \mathbb N^{M \times 3}$ denotes $M$ triangle faces composed by these vertices. And a motion $\mathbf m$ on one character is defined as a time sequence of pose meshes $\{(\mathbf V(t), \mathbf F )| t=0, 1, \cdots, \mathcal T\}$ of the character. Then, skeleton-free motion retargeting is to predict the most plausible target motion $\mathbf m_T$, given motion $\mathbf m_S$ and rest mesh $({\mathbf {\bar V}}_T,\mathbf F_T)$, i.e.,
    \begin{equation}
    \begin{aligned}
	\left \{(\mathbf V_S(t),\mathbf {\bar V}_S, \mathbf F_S )\right \}^{\mathcal T}_{t=0},({\mathbf {\bar V}}_T,\mathbf F_T ) 
 \mapsto \{ \mathbf {\tilde V}_T(t)\}_{t=0}^{\mathcal T}	
    \label{eq::obj}
    \end{aligned}
    \end{equation}
    where we use subscripts $S$ and $T$ to denote the source and the target characters, and $\mathbf {\bar V}$ to denote a rest mesh. Notice that pose retargeting~\cite{chen2021intrinsic,song20213d} is a special case of motion retargeting when $\mathcal T$ is set to $0$. To ensure our framework is applicable to both of the two tasks, we do not exploit inter-frame information to enhance the retargeting performance and use $\mathbf {V}$ to represent ${\mathbf V}(t)$ for short. An informative overview of this framework is illustrated in Figure~\ref{fig::pipe1}. 
    Given the deformation of source mesh $({\mathbf V}^{(0)}_S,{\mathbf {\bar V}}^{(0)}_S,{\mathbf F}_S^{(0)})$ and target rest mesh $({\mathbf V}_T^{(0)},{\mathbf F}_T^{(0)})$, a recursive mesh-coarsening process is applied to \hoe{coarsen input meshes for $K$ times to the lowest-resolution mesh possessing fewest vertices while keeping character morphology, i.e., $({\mathbf V}^{(K)}_S,{\mathbf {\bar V}}^{(K)}_S,{\mathbf F}_S^{(K)}),({\mathbf V}_T^{(K)},{\mathbf F}_T^{(K)})$. 
    Then, a retargeting layer is applied on the coarsest mesh to acquire a target pose mesh $\mathbf {\tilde V}_T^{(K)}$ that best preserves local-part motions. To improve the mesh quality of $\mathbf {\tilde V}_T^{(K)}$, a hierarchical refinement process is applied based on local deformation $({\mathbf {J}}_T^{(k)},{\mathbf W}_T^{(k)})$ learned from higher-resolution meshes.}
        
    \subsection{\hoe{Mesh-coarsening Module}\label{sec::met2}}
    \hoe{\textbf{To preserve inherent local-part motions}, we propose a mesh-coarsening module to coarsen the mesh representations for better motion transfer}.
    Reasonable skinning weights are crucial for comprehending the deformation process of a given mesh, which has been shown in the previous literature~\cite{li2021learning,liao2022skeleton,mosella2022skinningnet}. However, since the skeleton structure is strikingly varied with divergent stylized characters, predicting skinning weights in a unified way is challenging, especially when too many \hoe{interdependent and small body parts exist on the character mesh.}
    To address this issue, we revisit a quadratic-error-based coarsening (QEC) technique~\cite{garland1997surface} and equip each input HR mesh $({\mathbf {\bar V}}^{(k)},{\mathbf F}^{(k)})$ with an LR mesh $({\mathbf {\bar V}}^{(k+1)},{\mathbf F}^{(k+1)})$ recursively that possesses fewer vertices \hoe{while keeping similar character morphology.} We first rewrite the vertex representation as a 4D vector, i.e., $\mathbf v:=(v_x,v_y,v_z,1)$ to satisfy the error format. The objective is to find the minimal quadratic error with a given number of vertices $|{\mathbf V}^{(k+1)}|$, i.e.,
    \begin{equation}
    \begin{aligned}
	\!\arg\!\min_{\substack{{\mathbf {\hat V}}^{(k+1)}\\{\mathbf F}^{(k+1)}}}\!\sum_{{\mathbf v}({\mathbf v}_i,{\mathbf v}_j)}^{\mathbf {\hat V}^{(k+1)}}\!{\mathbf v}\!({\mathbf Q}_i\!+\!{\mathbf Q}_j)\!{\mathbf v'},\ \!|{\mathbf {\hat V}}^{(k+1)}|\!=\!|{\mathbf {V}}^{(k+1)}| 
    \label{eq::mc::1}
    \end{aligned}
    \end{equation}
    where $\hat V^{(k+1)}$ is a candidate vertex set composed by all valid contracted vertices from $V^{(k)}$, $\mathbf v \in \hat V^{(k+1)}$ is a contracted vertex from ${\mathbf v}_i$ and ${\mathbf v}_j$, ${\mathbf v}'$ is its transpose, ${\mathbf v}_i,{\mathbf v}_j \in {\mathbf V}^{(k)}$ are either comprising the edge set ${\mathbf E}^{(k)}$ or satisfying $\|{\mathbf v}_i - {\mathbf v}_j\|_2 < \epsilon$ ($epsilon$ is a small value with respect to the mesh scale), ${\mathbf Q}_i, {\mathbf Q}_j$ is heuristic quadratic metrics on vertices ${\mathbf v}_i, {\mathbf v}_j$, respectively. The above objective could be simply optimized by solving $|\mathbf V^{(k+1)}| \times 3$ linear problems, i.e.,
    \begin{equation}
    \begin{aligned}
        \frac{\partial {\mathbf v}({\mathbf Q}_i + {\mathbf Q}_j){\mathbf v}'}{\partial \mathbf v} = (0,0,0,1),\quad {\mathbf v} \in {\mathbf V}^{(k+1)}.
    \label{eq::mc::2}
    \end{aligned}
    \end{equation}
    Then we sort and contract the vertex with the largest error recursively from ${\mathbf {\hat V}}^{(k+1)}$ into ${\mathbf {V}}^{(k+1)}$ until the required number of vertices is satisfied, i.e., $|\mathbf{\hat V}^{(k+1)}| = |{\mathbf V}^{(k+1)}|$. 
    Detailed descriptions and analyses of this algorithm can be found in \cite{garland1997surface}.

    However, the resulting LR mesh shares no continuous mappings with the original HR mesh due to discretized vertex selections, which forbids us from deforming an LR mesh corresponding with the HR mesh. To solve this issue, we further compute two approximate assignment mappings ${\mathbf A}^{(k)},{\mathbf A}^{(k)}_{inv}$ between vertices of the two meshes, according to the barycentric coordinates~\cite{hille2002analytic} defined on their nearest faces. Thereinto,
    \begin{equation}
    \begin{aligned}
        {\mathbf A}_i :=& \{{\rm norm}({\mathcal B}({\mathbf v}_i,\mathbf f,{\mathbf V}^{(k)})) \\
        &| {\mathbf f} = \arg\min_{{\mathbf f}' \in {\mathbf F}^{(k)}} {\mathcal D}({\mathbf v_i},{\mathbf f}'), {\mathbf v}_i \in {\mathbf V}^{(k+1)} \}
    \label{eq::mc::3}
    \end{aligned}
    \end{equation}
    where $\mathcal D(\mathbf v,\mathbf f)$ measures the euclidean distance between $\mathbf v$ and $\mathbf f$, ${\mathcal B}({\mathbf v}_i, \mathbf f, {\mathbf V^{(k)}})$ outputs the barycentric coordinate of ${\mathbf v}_i$ with respect to vertices in $\mathbf f$. The sum of each row $i$ of $\mathbf A$ is normalized to $1$ to avoid scaling mesh.

    Notably, ${\mathbf A}^{(k)}_{inv}$ is usually not an inverse of ${\mathbf A}^{(k)}$ since the mapping to the nearest faces is injective, but we could compute it in the same way described above.

    In conclusion, we propose a mesh coarsening module based on the QEC algorithm \hoe{to address issues on preserving local-part motions. It produces morphologically similar LR meshes with the input mesh and establishes their deforming relations by giving two efficient continuous mappings.} Experiments in Section~\ref{sec::exp::verf::1} (a) and (b) further verify its effectiveness in preserving \hoe{the morphology and small-part motions} of input character meshes, respectively.
    The inverse mapping ${\mathbf A}^{(k)}_{inv}$ will be involved in the next section.
    
\begin{figure*}[t]
    \centering
    \includegraphics[width=1.9\columnwidth]{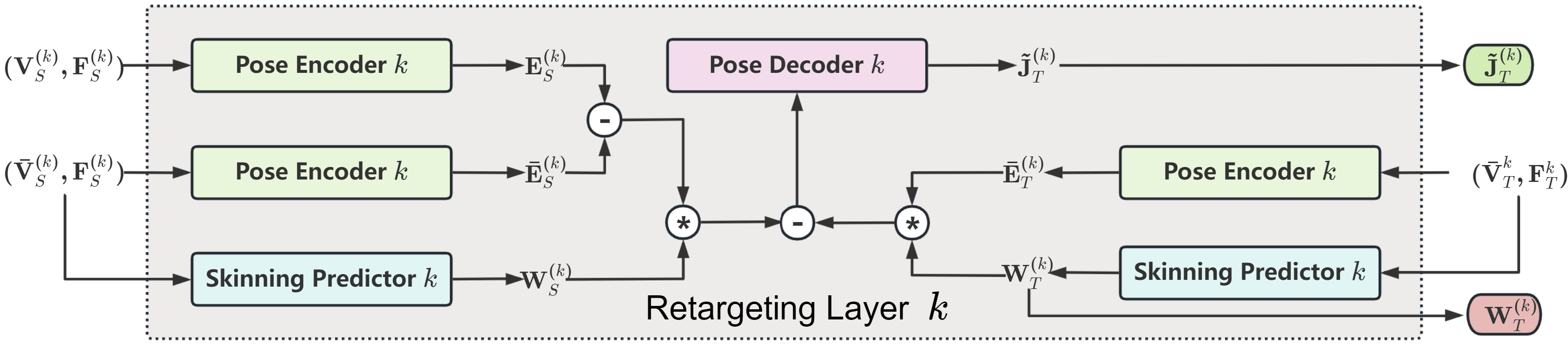}
    \caption{Design of retargeting process at every resolution. Each retargeting layer contains three modules, including Pose Encoder, Pose Decoder, and Skinning Predictor. Thereinto, Pose Encoder captures vertex-wise pose information (i.e., $\mathbf E^k_S$ and $\mathbf {\bar E}^k_S$) of a given mesh, while Skinning Predictor predicts potential skinning weights of each vertex, paving the way for aggregating part-level pose embedding. Finally, Pose Decoder decodes the deformation of each part $\mathbf{\tilde J}^k_T$ related to the skinning weights of a target mesh $\mathbf W^k_T$.}
    \label{fig::pipe2}
\end{figure*}

    \subsection{\hoe{Hierarchical Refinement Procedure}\label{sec::met3}}
    \hoe{\textbf{To produce high-quality pose meshes}, we leverage a hierarchical refinement procedure to complement missing mesh details with LR meshes.}
    Inspired by SfPT~\cite{li2021learning}, we adopt a \hoe{similar retargeting pipeline on higher-resolution meshes} that first predicts skinning weights on source and target rest meshes and then learns part-wise mesh deformations \hoe{to refine the previous LR mesh output.}
    
    As depicted in Figure~\ref{fig::pipe2}, a retargeting layer contains three modules, including Pose Encoder, Pose Decoder, and Skinning Predictor. Thereinto, Pose Encoder is a three-layer graph convolution network that captures vertex-wise pose information (i.e., ${\mathbf E}^{(k)}_S$, ${\mathbf {\bar E}}^{(k)}_S$, and ${\mathbf {\bar E}}^{(k)}_T$) of given meshes. Meanwhile, Skinning Predictor possesses the same structure as Pose Encoder but predicts the potential skinning weights of each vertex, i.e., $\mathbf W \in \mathbb R^{J \times V}$ ($J$ is the number of body parts for deformation), paving the way for aggregating part-level motion representations. Then, a part-wise mesh deformation of the target character could be represented as
    \begin{equation}
    \begin{aligned}
        {{\mathbf W}_S^{(k)}}({\mathbf E}_S^{(k)} - {\mathbf {\bar E}}_S^{(k)}) - {\mathbf W}^{(k)}_T{\mathbf {\bar E}}_T^{(k)} 
    \label{eq::rl::1}
    \end{aligned}
    \end{equation}
    where ${{\mathbf W}_S^{(k)}}({\mathbf E}_S^{(k)} - {\mathbf {\bar E}}_S^{(k)})$ embeds character-independent deformation at part levels, and ${\mathbf W}^{(k)}_T{\mathbf {\bar E}}_T^{(k)}$ indicates the initial state of target mesh ready to deform. 

    Before refining LR meshes, we first model unified part-level transformations ${\mathbf J} \in \mathbb R^{J \times 6}$ from the rest meshes to pose meshes as a concatenation of joint positions ${\mathbf J}_{pos}$ and joint rotations ${\mathbf J}_{rot}$. Both of them could be estimated by vertex transformations using linear blend skinning~\cite{kavan2014direct}, i.e.,
    \begin{equation}
    \begin{aligned}
        {\mathbf V}_i &= \sum_{j=1}^J W_{j,i} {\mathbf J}_{rot,j} ({\mathbf {\bar V}}_i - {\mathbf J}_{pos,j}), \\
        {\mathbf J}_{pos,j} &= {\mathbf W}_j{\mathbf {\bar V}} \|{\mathbf W}_j\|^{-1}_1.
    \label{eq::rl::2}
    \end{aligned}
    \end{equation}
    



    \hoe{Then, our idea is to learn local deformations on Pose Decoder from HR meshes and apply these deformations to refine the LR mesh output.}
    As we have computed an inverse mapping ${\mathbf A}^{(k)}_{T,inv} \in \mathbb R^{V^{(k)}_T \times V^{(k+1)}_T}$, the learning objective of each retargeting layer can be simply interpreted as
    \begin{equation}
    \begin{aligned}
        &\arg\min_{ {\mathbf J}^{(k)}_{T}, {\mathbf W}^{(k)}_T}
        \|\Delta {\mathbf {V}}^{(k)}_T\|_2 \\
        \Leftrightarrow & \arg\min_{ {\mathbf J}^{(k)}_{T}, {\mathbf W}^{(k)}_T}
        \|{\mathbf V}_T^{(k)} - {\mathbf A}^{(k)}_{T,inv}{\mathbf {\tilde V}}_T^{(k+1)}\|_2
    \label{eq::rl::3}
    \end{aligned}
    \end{equation}
    , where $\Delta {\mathbf {V}}^{(k)}_T$ is \hoe{the local deformation at layer $k$}, and ${\mathbf V}_T^{(k)}$ is the corresponding ground-truth pose mesh. 
    
    Practically, this objective is backbreaking to learn since ${\mathbf A}^{(k)}_{T,inv}{\mathbf {\tilde V}}_T^{(k+1)}$ intrinsically neglects details on the rest mesh due to the matching between vertices. It means \hoe{each HR local deformation} needs to simultaneously \hoe{refine} the HR rest pose of a target character and deform incremental details based on this pose.
    
    To ease this burden, we further propose an approximate objective $\Delta {\mathbf {\tilde V}}^{(k)}_T$ that complements the resolution gap between LR and HR rest meshes at the body-part level, i.e.,
    \begin{equation}
    \begin{aligned}
    &\Delta {\mathbf {V}}^{(k)}_T + {\mathbf A}^{(k)}_{T,inv}{\mathbf {\tilde V}}_T^{(k+1)}
    \approx \Delta {\mathbf {\tilde V}}^{(k)}_T\\
    =& {\rm LBS}({\mathbf J}_T^{(k,\cdots,K)}, {\mathbf W}_T^{(k,\cdots,K)})
    \label{eq::rl::4}
    \end{aligned}
    \end{equation}
    where ${\rm LBS}(\cdot,\cdot)$ is an LBS-like operator that outputs a pose mesh given rigid transformations $\mathbf J$ and skinning weights $\mathbf W$, and (${\mathbf J}_T^{(k,\cdots,K)},{\mathbf W}_T^{(k,\cdots,K)})$ denotes a part-level representation that is leveraged to deform a $(k+1)$-th coarse mesh, defined in a recursive format, i.e.,
    \begin{equation}
    \begin{aligned}
        {\mathbf J}_T^{(k,\cdots,K)} &:= \left [\mathbf J_T^{(k)}, {\mathbf J}_T^{(k+1,\cdots,K)} \right ]\\
        {\mathbf W}_T^{(k,\cdots,K)} &:= \left [ \mathbf W_T^{(k)}, {\mathbf A}^{(k)}_{T,inv} {\mathbf W}_T^{(k+1,\cdots,K)} \right ]
       \label{eq::rl::5} 
    \end{aligned}
    \end{equation}
    where 
    $[\cdots]$ means a concatenation of tensors at the zero dimension. 
    \hoe{By leveraging $\Delta {\mathbf {\tilde V}}^{(k)}_T$, we assume the initial states of HR vertices are known and lead the model to concentrate on refining local deformation only. }
    
    To summarize, we proposed in this section \hoe{a hierarchical refinement procedure that complements missing details on LR meshes based on a reasonable approximate objective.}
    Experiments in Section~\ref{sec::exp::verf::3} (c) further show its efficiency in \hoe{improving mesh quality} in contrast with naive strategies.
    
    \subsection{Training \& Losses\label{sec::met4}}
    \subsubsection{Training Strategies}
    Our proposed model could be trained both with paired and unpaired motion data. For paired data, predicted target meshes $\{{\mathbf {\tilde V}}_T^{(k)}\}^{K}_{k=0}$ with different resolutions are supervised by the corresponding coarsened ground-truth meshes $\{{\mathbf V}_T^{(k)}\}^{K}_{k=0}$. For unpaired data, since ${\mathbf V}^{(0)}_T$ is agnostic, we apply a cycle consistency loss~\cite{zhu2017unpaired,aberman2020skeleton,liao2022skeleton} to enhance cross-domain performance,i.e., given a retargeting model $\mathcal M$ and the source and target characters $S$ and $T$,
    \begin{equation}
    \begin{aligned}
        &\mathcal L_{cyc}(S,T) := \sum_{(A,B)}^{\{(S,T),(T,S)\}} \sum_{k=0}^{K} \\
        &\quad \| {\mathbf {V}}^{(k)}_{A} - 
         \mathcal M( \mathcal M ( {\mathbf {V}}^{(k)}_{A} , {\mathbf {\bar V}}^{(k)}_{A}, {\mathbf {\bar V}}^{(k)}_{B}) ,{\mathbf {\bar V}}^{(k)}_{B}, {\mathbf {\bar V}}^{(k)}_{A})\|_1
    \label{eq::tl::cyc}
    \end{aligned}
    \end{equation}
    where $\mathcal M ( {\mathbf {V}}^{(k)}_{A} , {\mathbf {\bar V}}^{(k)}_{A}, {\mathbf {\bar V}}^{(k)}_{B})$ denotes a retargeting process from $A$ to $B$. We omit their faces $\mathbf F^{(k)}_{A}$ and $\mathbf F^{(k)}_{B}$  here for simplicity.

    Besides, as we observe that learning on LR meshes is likely to preserve motion semantics of mesh deformation (see Section~\ref{fig::exp::verf::2} (b)), which benefits the subsequent learning on mesh details, faster converging on retargeting LR meshes is desirable for better performance. 
    To achieve this, we introduce varied loss weights on \hoe{HR and LR layers} at different training epochs, i.e.,
    \begin{equation}
    \begin{aligned}
        \mathcal L_{ret}(x)\!:=\!\left[\!1-\sum_{k=1}^{K}\!w_x(\alpha_k,\beta_k)\!\right] \!L^{(0)}_{ret}\!+\!\sum_{k=1}^{K}\!w_x(\alpha_k,\beta_k)\mathcal L^{(k)}_{ret}
    \label{eq::tl::var1}
    \end{aligned}
    \end{equation}
    where $w(\alpha_k,\beta_k,x)$ linearly interpolates between $[\alpha_k,\beta_k]$ regarding the $x$-th epoch, i.e., $x \in [0,x_{max}]$
    \begin{equation}
    \begin{aligned}
        w_x(\alpha_k,\beta_k)\!=\!\alpha_k\!+\!\frac{x}{x_{max}}(\beta_k - \alpha_k)
        ,\ \sum_{k=1}^{K} w_x(\alpha_k,\beta_k) < 1.
    \label{eq::tl::var2}
    \end{aligned}
    \end{equation}
    By setting decreasing weights on LR meshes, the model concentrates on \hoe{learning better motion transfer on target characters} at first, and is gradually fine-tuned when more and more high-frequency details on HR meshes are considered. We further verify the improvement of this strategy against fixed loss weights in Section~\ref{sec::exp::abl}.
    \subsubsection{Base Losses}
    In addition to the strategies discussed above, we describe here the other losses helpful to improve retargeting performance.

    \textbf{Layer Retargeting Loss ${\mathcal L}^{(k)}_{ret}$} This loss basically ensures that retargeted meshes are similar to the ground-truth meshes both in part and vertex levels, i.e.,
    \begin{equation}
    \begin{aligned}
        {\mathcal L}^{(k)}_{ret} = \|{\mathbf {V}}_T^{(k)} - {\mathbf {\tilde V}}_T^{(k)}\|_1 + \|{\mathbf {J}}_T^{(k)} - {\mathbf {\tilde J}}_T^{(k)}\|_1
    \label{eq::tl::ret}
    \end{aligned}
    \end{equation}
    
    \textbf{Skinning Similarity Loss ${\mathcal L}_{sk}$} This loss, comprising intra-layer similarity and inter-layer similarity losses (i.e., ${\mathcal L}_{intra}$ and ${\mathcal L}_{inter}$), ensures body parts are divided similarly between the HR and LR meshes, the source and target meshes, the predicted and ground-truth meshes (if ground-truth skinning weights of characters are given).
    \begin{equation}
    \begin{aligned}
        &{\mathcal L}^{(k)}_{sk} = \sum_{k=0}^K {\mathcal L}^{(k)}_{inter} + \sum_{k=1}^K {\mathcal L}^{(k),(k-1)}_{intra}\\
        &{\mathcal L}^{(k)}_{inter} =\!\sum_{j=1}^{J}{\rm KL}({\mathbf W}_{T,j}^{(k)}\|{\mathbf W}^{(k)}_{S,j})\!+\!\sum^{\{S,T\}}_A\!{\rm KL}(\!{\mathbf 
        W}_{A,j}^{(k)}\|{\mathbf {\bar W}}^{(k)}_{A,j}\!)\!\\
        &{\mathcal L}^{(k),(k-1)}_{intra} = \sum_{j=1}^{J}
        {\rm KL}({\mathbf W}_{T,j}^{(k)}\|{\mathbf W}^{(k-1)}_{S,j})
    \label{eq::tl::ret}
    \end{aligned}
    \end{equation}
    where ${\rm KL}({\mathbf p}\|{\mathbf q})$ denotes Kullback-Leibler divergence that measures the similarity between probabilities $\mathbf p$ and $\mathbf q$, 
    and ${\mathbf {\bar W}}^{(k)}_{A,j}$ is the skinning weights of a selected ground-truth joint that is most similar to ${\mathbf {\bar W}}^{(k)}_{A,j}$ (since the number of joints usually dis-matches the number of body parts in our model.)

    \textbf{Rigid Loss ${\mathcal L}_{rig}$} This loss ensures a rigid transformation of vertices with their neighbors, i.e.,
    \begin{equation}
    \begin{aligned}
        {\mathcal L}_{rig}\!=\!\sum_{k=0}^{K}\!\sum_{({\mathbf v}_a,{\mathbf v}_b)  \in {\mathbf V}^{(k)}_T}\!\bigg| \| {\mathbf {\tilde v}}_a - {\mathbf {\tilde v}}_b \|_2  - \| {\mathbf {\bar v}}_a - {\mathbf {\bar v}}_b \|_2 \bigg |
    \label{eq::tl::ret}
    \end{aligned}
    \end{equation}

    
\section{Experiments~\label{sec::exp}}
We show in this section the comprehensive performance of the proposed method. Specifically, We initially describe the evaluating datasets and parameter setting of our model (See Section~\ref{sec::exp::data}). Then, extensive quantitative and visualization results are presented in Section~\ref{sec::exp::quan} and Section~\ref{sec::exp::visual}, respectively. In addition, we verify the assumptions made in the method section (See Section~\ref{sec::exp::verf}) and convey an ablation study on the proposed techniques in Section~\ref{sec::exp::abl}.  
\subsection{Datasets \& Parameter Setting~\label{sec::exp::data}}
Following \cite{liao2022skeleton}, our model is trained and evaluated on three renowned 3D character datasets. 

\textbf{AMASS~\cite{mahmood2019amass} }is a human motion capture dataset based on Skinned Multi-Person Linear Model (SMPL)~\cite{loper2015smpl}. Since SMPL describes human mesh by disentangled shape and pose parameters, paired data of different characters are available. We randomly sample $115622$ poses on some frames of the original motion sequences as our dataset ($101085$ poses for train, and $14537$ poses for test)

\textbf{Mixamo~\cite{mixamo2023} }is an online animation website that contains hundreds of humanoid characters and different kinds of motions to animate them. Like AMASS, paired motion data is available by rotation copy, owing to a unified skeleton template embedded in each character. We utilize $1112$ motions with various styles and $100$ characters in our experiments ($778$ motions and $70$ for train, and $334$ motions and $30$ characters for test).

\textbf{RigNet~\cite{xu2020rignet} }is a diversified character dataset comprising highly-stylized humanoid and non-humanoid characters. Only the rest meshes of each character and their skeletal rigs are available. We cannot construct paired motion data since each character possesses a specific heterogeneous skeleton structure where rotation copy and interpolation are not supported. We select $1888$ and $810$ characters for training and test, respectively.

The parameter settings of our model are stated as follows. We mix the above three datasets at the rates of $(0.25,0.5,0.25)$ as our training data and use two different resolution layers to reach the best performance balanced on time and quality. The dimension of motion representations at each frame $\mathbf E$ is set to $128$, and each retargeting layer outputs $40$ parts to deform target meshes. The varying weights of retargeting loss on the coarse mesh are set to $(0.6,0.4)$ (i.e., $\alpha_1=0.6, \beta_1=0.4$, according to Equation~\ref{eq::tl::var2}).

\subsection{Baseline methods~\label{sec::exp::baseline}}
We compared our method with the previous SOTA methods regarding skeleton-free motion retargeting.

\textbf {SfPT~\cite{liao2022skeleton} }is the first work that automatically transfers poses or motions between input meshes without any skeletal rigging or skinning. The idea is to predict skinning weights on each character and transfer the part-wise rigid transformations between the source and target characters. We additionally compare our method with a variant of SfPT (i.e., SfPT-GT) which directly applies ground-truth source transformations on the target mesh.

\textbf {NBS~\cite{li2021learning} }is the SOTA method that animates rest meshes with input skeleton motion sequence. It first envelops a given skeleton template into the target mesh by predicting joint positions as well as skinning weights, and then motions could be retargeted by a direct copy of joint rotations. 

\subsection{Quantitative Results~\label{sec::exp::quan}}
We quantitatively evaluate our method by point-wise mesh euclidean distance (PMD)~\cite{wang2020neural,zhou2020unsupervised} on mesh reconstruction and retargeting tasks. For AMASS and Mixamo datasets, we measure predicted meshes with the ground-truth pose meshes. For RigNet dataset without ground-truth poses, we report the cycle error to reconstruct the source meshes (i.e., evaluating a retargeting chain $S \rightarrow T \rightarrow \tilde S$.).
\begin{table}[t]
    \begin{center}
    \smallskip
    \resizebox{1.0\columnwidth}{!}
    {
        \setlength{\tabcolsep}{2.4mm}{ 
            \begin{tabular}{ c | c c | c c | c c}
                \toprule[1.5pt]
                \multirow{2}{*}{\textbf{Method}} & \multicolumn{2}{c|}{\textbf{AMASS ($10^{-4}$)}} & \multicolumn{2}{c|}{\textbf{Mixamo ($10^{-2}$)}} & \multicolumn{2}{c}{\textbf{RigNet ($10^{-2}$)}}\\
                \cline{2-7}
                & \textbf{ret.}$_{\downarrow}$ & \textbf{rec.}$_{\downarrow}$ & \textbf{ret.}$_{\downarrow}$ & \textbf{rec.}$_{\downarrow}$ & \textbf{cyc.}$_{\downarrow}$ & \textbf{rec.}$_{\downarrow}$ \\
                \midrule[0.5pt]
                NBS& $5.052$& $4.806$& $5.711$& $3.211$& - & -\\
                SfPT-GT&        $5.100$&    -   & $9.894$& -      & $2.758$& -\\
                SfPT&           $5.148$& $4.924$& $4.029$& $2.285$& $6.700$& $4.154$\\
                \midrule[0.5pt]
                \textbf{ours.}& $\textbf{4.569}$& $\textbf{4.359}$& $\textbf{2.649}$& $\textbf{0.817}$& $\textbf{1.907}$& $\textbf{0.400}$\\
                \bottomrule[1.5pt]
            \end{tabular}}
        }
    \end{center}
    \caption{Quantitative comparisons among different methods. \textbf{rec.: }a self-retargeting that reconstructs source pose meshes. \textbf{ret.: }a normal retargeting between source and target characters. \textbf{cyc.: }a cycling-retargeting that reconstructs source pose meshes.}
    \label{tab::quan}
\end{table}
We see from Table~\ref{tab::quan} that our method outperforms the existing methods on the given tasks. This is because, with \hoe{mesh-coarsening and hierarchical refinement processes} on the target mesh, the model learns to \hoe{perform well on two disentangled objectives, i.e., better motion transfer and better mesh quality}, which improves the performance on both retargeting and reconstruction tasks. Besides, it also reveals that SfPT-GT outperforms SfPT in two datasets (i.e., AMASS and RigNet), which indicates the transformations on source characters are generally good enough to deform target meshes, and SfPT sometimes may fail to learn better target transformations regarding the source ones, due to inflexible and insufficient part-wise correspondence learning among input characters.

\begin{figure*}[t]
    \centering
    \includegraphics[width=1.9\columnwidth]{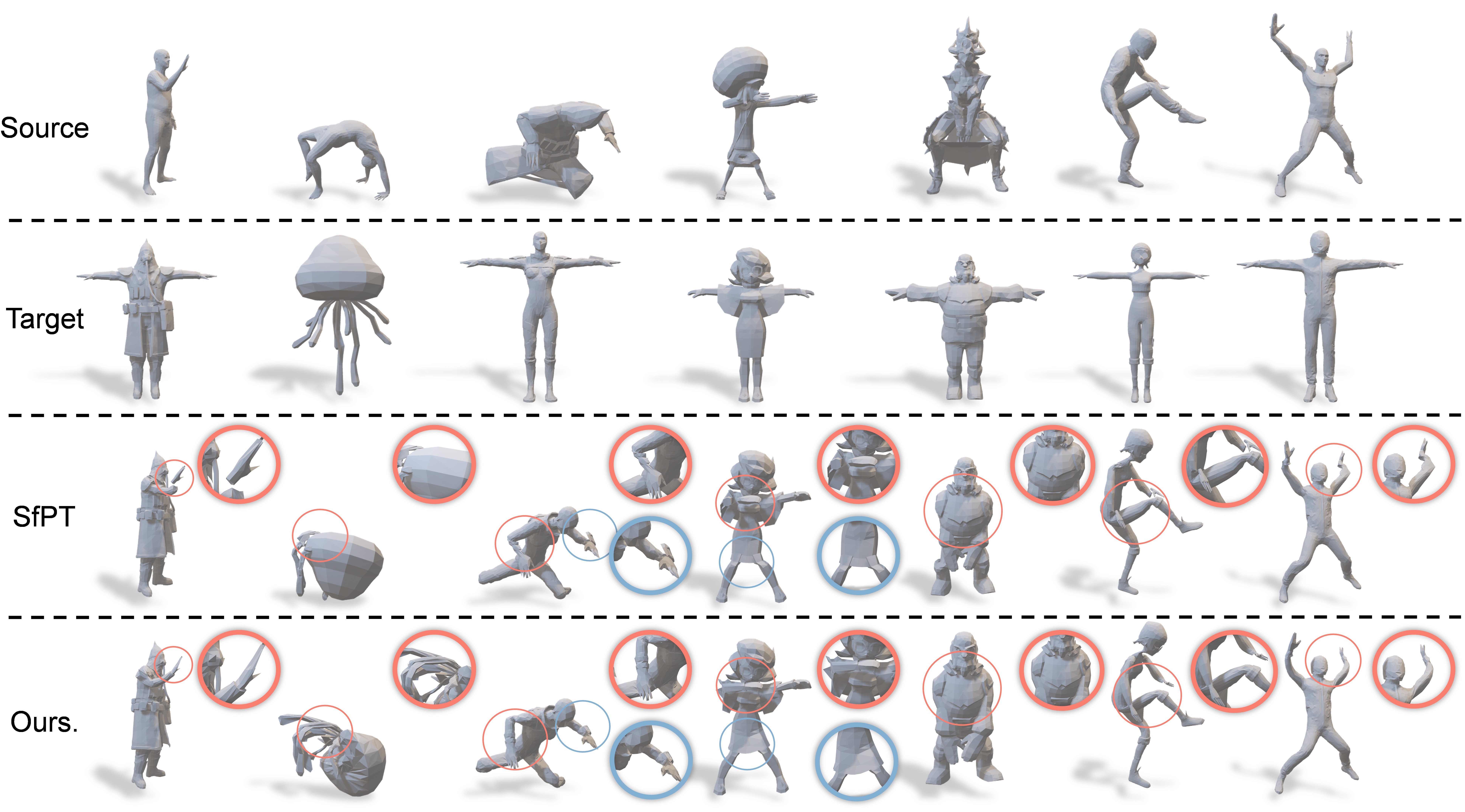}
    \caption{Visualization results of motion and pose retargeting. \textbf{Columns from left to right} are different pairs of characters for motion or pose retargeting.}
    \label{fig::exp::visual}
\end{figure*}

\subsection{Visualization Results~\label{sec::exp::visual}}
We visually compare our method with the SOTA method (i.e., SfPT) on motion and pose retargeting tasks. Thereinto, we retarget poses from AMASS dataset by selecting a group of test poses and characters and retarget motions from Mixamo characters. The retargeting results on highly-stylized characters both in Mixamo and RigNet datasets are depicted in Figure~\ref{fig::exp::visual} 
(see more visualization results and animated videos at \url{https://semanticdh.github.io/HMC/}).
We observe that our model outputs more plausible target meshes regarding the source deformation, regarding both \hoe{small-part motions and local motion interdependence.} Specifically, for \hoe{\emph{local motion interdependence}}, we see many failure cases occurring in SfPT are 
fixed by our model (See fractured right hand in Column~$1$, warped right forearm in Column~$3$, penetrated legs and right hands in Column~$4$, penetrated arms in Column~$5$, fractured hands in Column~$5$). 
For \hoe{\emph{handling small-part motions}}, our model also performs better. 
In Column~$3$, as illustrated in the target rest pose, the watch accessory on the left hand appears to be attached to the hand. It shows that our model preserves \hoe{their attaching relationship} well in the retargeted pose mesh. Similar cases can also be found in Columns~$2$, i.e., keeping the tentacles of a jellyfish attached to the bottom of its head. Besides, the left hand in Column~$6$ also reveals that our model \hoe{accurately outputs hand movement from the source}, i.e., Raising the left hand straight. Through these encouraging examples, we conclude that the \hoe{mesh-coarsening} of input meshes does provide \hoe{beneficial} information for comprehending the retargeting process, \hoe{especially on preserving local-part motions.} 



\subsection{Verification Results~\label{sec::exp::verf}}
We verify the assumptions in the proposed techniques by giving sufficient experimental evidence.

\begin{figure}[t]
    \centering
    \includegraphics[width=0.8\columnwidth]{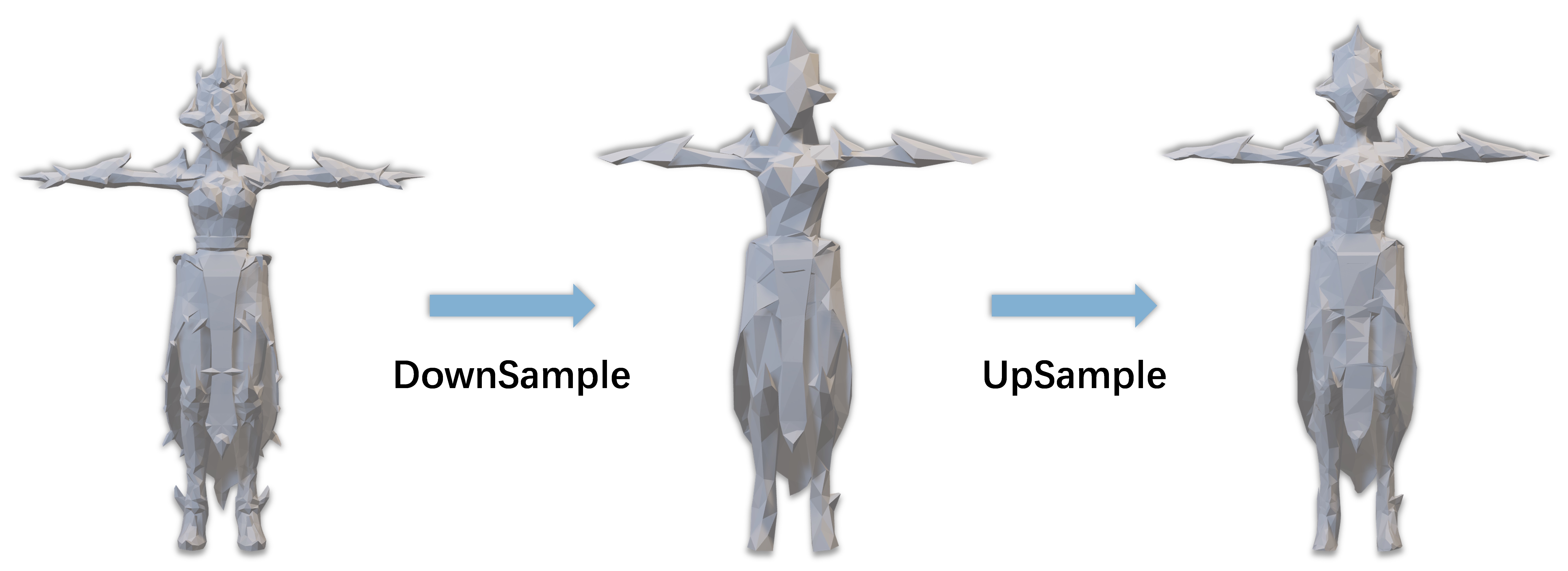}
    \caption{\hoe{Performance of the mesh-coarsening module.} \textbf{Left: }input mesh. \textbf{Middle: } coarsened mesh with $0.6^3$ vertices. \textbf{Right:} \hoe{upsampled mesh by leveraging ${\mathbf A}^{\cdot}_{T,inv}$.} }
    \label{fig::exp::verf::1}
\end{figure}

\textbf{(a) \hoe{performance of the mesh-coarsening module}~\label{sec::exp::verf::1} }In Section~\ref{sec::met1}, we conclude two continuous mappings between HR and LR meshes for deformation. Here, we leverage this technique to \hoe{recursively coarsen} meshes three times from the input mesh at a coarsening ratio of $0.6$ and then up-sample from the lowest resolution mesh to reconstruct the original mesh. Figure~\ref{fig::exp::verf::1} illustrates the two processes, reflecting that both of the processes well \hoe{preserve the character morphology}. 

\begin{figure}[t]
    \centering
    \includegraphics[width=0.8\columnwidth]{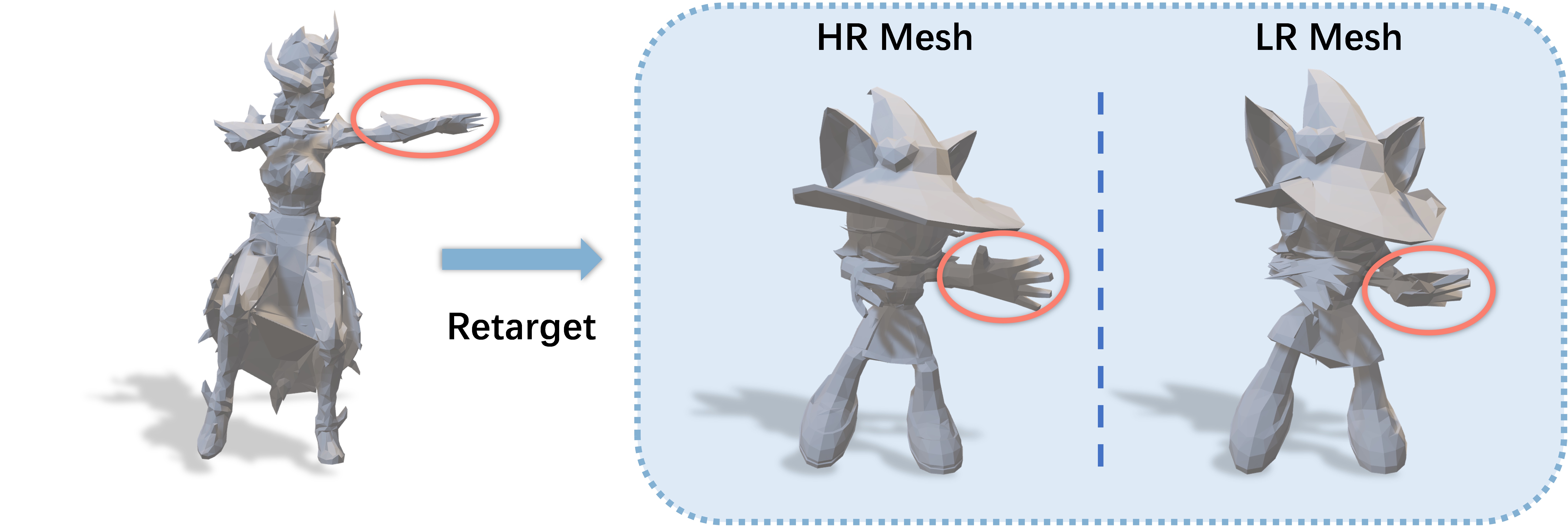}
    \caption{\hoe{Preservation of small-part motions} by LR-mesh retargeting. \textbf{Left: }input source motion. \textbf{Middle: }retarget to an HR target mesh using a retargeting model trained by HR meshes. \textbf{Right: }retarget to an LR target mesh using a retargeting model trained by LR meshes.}
    \label{fig::exp::verf::2}
\end{figure}

\textbf{(b) preservation of \hoe{small-part motions}~\label{sec::exp::verf::2} }By using a coarse mesh with \hoe{reduced and rebalanced} vertices, motion transfer is easier to learn, as stated in Section~\ref{sec::met1}. Here We train a retargeting model with pure LR meshes (the coarsening ratio is set to $0.6$) and compare the retargeting results with that trained with pure HR meshes. Figure~\ref{fig::exp::verf::2} reveals the superior performance of LR-mesh retargeting. We see the left hand is parallel to the ground like the ground truth when inputting an LR mesh. 

\begin{figure}[t]
    \centering
    \includegraphics[width=0.8\columnwidth]{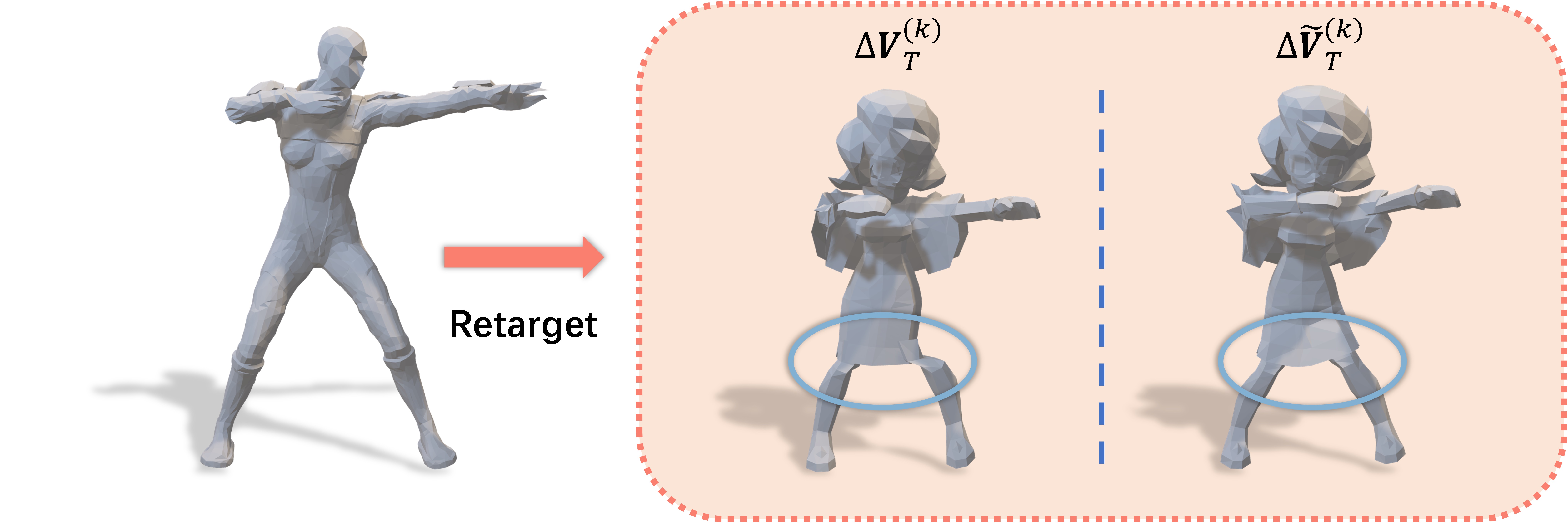}
    \caption{Comparison regarding the \hoe{refining objectives} between our proposed $\Delta {\mathbf {\tilde V}}_{T}^{(k)}$ and a naive objective $\Delta {\mathbf V}_{T}^{(k)}$.}
    \label{fig::exp::verf::3}
\end{figure}

\textbf{(c) refinement process on the approximate objective~\label{sec::exp::verf::3} }As we propose an approximate objective $\Delta {\mathbf {\tilde V}}^{(k)}_T$ instead of $\Delta {\mathbf {V}}^{(k)}_T$ to \hoe{avoid learning the initial states of HR mesh vertices}, here we show in Figure~\ref{fig::exp::verf::3} the differences when using either of the two objectives trains the model. The results indicate the necessity of easing the burden of learning redundant details on HR rest meshes.

\begin{table}[t]
    \begin{center}
    \smallskip
    \resizebox{1.0\columnwidth}{!}
    {
        \setlength{\tabcolsep}{2.4mm}{ 
            \begin{tabular}{ c | c c | c c | c c}
                \toprule[1.5pt]
                \multirow{2}{*}{\textbf{Method}} & \multicolumn{2}{c|}{\textbf{AMASS ($10^{-4}$)}} & \multicolumn{2}{c|}{\textbf{Mixamo ($10^{-2}$)}} & \multicolumn{2}{c}{\textbf{RigNet ($10^{-2}$)}}\\
                \cline{2-7}
                & \textbf{ret.}$_{\downarrow}$ & \textbf{rec.}$_{\downarrow}$ & \textbf{ret.}$_{\downarrow}$ & \textbf{rec.}$_{\downarrow}$ & \textbf{cyc.}$_{\downarrow}$ & \textbf{rec.}$_{\downarrow}$ \\
                \midrule[0.5pt]
                \textbf{w/o HR}& $9.384$& $9.255$& $\textbf{2.099}$& $\textbf{1.138}$& $\textbf{2.871}$&$\textbf{0.945}$\\
                \textbf{w/o ref}& $31.77$& $31.55$& $6.474$& $2.583$& $7.277$& $2.276$\\
                \textbf{w/o var}& $\textbf{3.005}$& $\textbf{2.916}$& $3.273$& $4.886$& $16.29$&$12.50$\\
                \midrule[0.5pt]
                \textbf{full approach.}&$\textbf{4.569}$& $\textbf{4.359}$& $\textbf{2.649}$& $\textbf{0.817}$& $\textbf{1.907}$& $\textbf{0.400}$\\
                \bottomrule[1.5pt]
            \end{tabular}}
        }
    \end{center}
    \caption{Ablation results on components in our model.}
    \label{tab::abl}
\end{table}

\subsection{Ablation Study~\label{sec::exp::abl}}
We investigate the effectiveness of crucial components in our model. i) \textbf{w/o HR} means the model is \hoe{purely trained on the LR mesh}. ii) \textbf{w/o ref} means the model directly mixes the outputs of each retargeting layer for prediction instead of \hoe{refining the LR mesh output by HR local deformations}. iii) \textbf{w/o var} means the varied loss weights on LR meshes defined by Equation~\ref{eq::tl::var1} and~\ref{eq::tl::var2} are deactivated (equivalent to set $\alpha_1=\beta_1=0.4$).
Table~\ref{tab::abl} shows the performance of these variants, indicating our full approach achieves the overall best performance by integrating these components. Specifically, we see without HR meshes (i.e., \textbf{w/o HR}), the model performs poorly on AMASS dataset due to the absence of perceiving subtle morphological differences in human shape. Besides, without varied loss weights (i.e., \textbf{w/o var}), the model overfits bias on given shapes and fails to predict poses on highly-stylized characters in RigNet dataset.

\section{Conclusion}
In this paper, we propose a \hoe{hierarchical mesh-coarsening method for skeleton-free motion retargeting}, and discuss two effective techniques to \hoe{preserve inherent local-part motions, including handling small-part motions and preserving local motion interdependence}. Extensive quantitative and qualitative experiments highlight the superior performance of our proposed method. 

\textbf{Limitations }Due to the lack of animated highly-stylized characters, our model still fails when a target character possesses \hoe{totally different morphology} from the source one (like quadrupeds from bipeds), despite most of the failure cases being fixed. We leave it for our future work.


{\small
\bibliographystyle{ieee_fullname}
\bibliography{sk-free_moret}
}

\end{document}